\documentclass[conference]{IEEEtran}

\usepackage[cmex10]{amsmath}
\usepackage{algorithmic}
\usepackage{url}
\usepackage{array}
\usepackage[pdftex]{graphicx}
\usepackage{commath}
\usepackage{svg}
\usepackage{amsmath}
\usepackage{mdwtab}
\usepackage{eqparbox}
\usepackage[tight,footnotesize]{subfigure}
\usepackage[utf8]{inputenc}
\usepackage{stfloats}
\usepackage{lipsum}
\usepackage{outlines}
\usepackage[acronym]{glossaries}
\usepackage{tabularx}
\usepackage{multirow}
\usepackage{balance}
\usepackage{pgfplots}

\usepackage{listings}
\usepackage{xcolor}

\definecolor{codegreen}{rgb}{0,0.6,0}
\definecolor{codeblue}{rgb}{0,0.0,4}

\lstdefinestyle{mystyle}{
	stringstyle=\color{codeblue},
    keywordstyle=\color{codegreen},
    basicstyle=\ttfamily\small,
    breakatwhitespace=false,         
    breaklines=true,                 
    captionpos=b,                    
    keepspaces=true,                 
    showspaces=false,                
    showstringspaces=false,
    showtabs=false,                  
    tabsize=2
}

\usepackage{csquotes} 

\usepackage[noadjust]{cite}

\usepackage[resetlabels,labeled]{multibib}
\newcites{S}{Survey References}

\AtBeginEnvironment{listing}{\setcounter{listing}{\value{lstlisting}}}
\AtEndEnvironment{listing}{\stepcounter{lstlisting}}

\definecolor{darkCyan}{RGB}{0,145,207}

\newacronym{cpps}{CPPS}{Cyber-Physical Production System}
\newacronym{etfa}{ETFA}{\emph{Emerging Technologies and Factory Automation}}
\newacronym{slr}{SLR}{Systematic Literature Review}
\newacronym{ppr}{PPR}{Product-Process-Resource}
\newacronym{owl}{OWL}{Web Ontology Language}
\newacronym{emf}{EMF}{Eclipse Modeling Framework}
\newacronym{bpmn}{BPMN}{Business Process Model and Notation}
\newacronym{isa}{ISA}{International Society of Automation}
\newacronym{vdi}{VDI}{Verein Deutscher Ingenieure}
\newacronym{vdma}{VDMA}{Verband Deutscher Maschinen- und Anlagenbau}
\newacronym{uml}{UML}{Unified Modeling Language}
\newacronym{plc}{PLC}{Programmable logic controller}
\newacronym{scada}{SCADA}{Supervisory control and data acquisition}

\begin{document}

\title{Capabilities and Skills in Manufacturing: \\A Survey Over the Last Decade of ETFA}

\author{
    \IEEEauthorblockN{
        Roman Froschauer\IEEEauthorrefmark{1},
        Aljosha Köcher\IEEEauthorrefmark{2},
        Kristof Meixner\IEEEauthorrefmark{3}\IEEEauthorrefmark{5},
        Siwara Schmitt\IEEEauthorrefmark{4}
        and~Fabian Spitzer\IEEEauthorrefmark{1}
    }
    
    \IEEEauthorblockA{
        \IEEEauthorrefmark{3}Christian-Doppler-Laboratory SQI, Information and Software Engineering\\
        TU Wien, Vienna, Austria
        \IEEEauthorrefmark{5}\emph{corresponding author}\\
        {\tt\small kristof.meixner@tuwien.ac.at}\\
        \IEEEauthorrefmark{2}Institute of Automation\\
        Helmut Schmidt University, Hamburg, Germany\\
        {\tt\small aljosha.koecher@hsu-hh.de}\\
        \IEEEauthorrefmark{1}Center of Excellence for Smart Production\\
        University of Applied Sciences Upper Austria, Wels, Austria\\
        {\tt\small [first].[last]@fh-wels.at}\\
        \IEEEauthorrefmark{4}Embedded Systems Engineering\\
        Fraunhofer IESE, Kaiserslautern, Germany\\
        {\tt\small siwara.schmitt@iese.fraunhofer.de}\\
    }
}

\maketitle

\begin{abstract}
Industry 4.0 envisions \glspl{cpps} to foster adaptive production of mass-customizable products. Manufacturing approaches based on capabilities and skills
aim to support this adaptability by encapsulating machine functions and decoupling them from specific production processes.
At the 2022 IEEE conference on \gls{etfa}, a special session on capability- and skill-based manufacturing is hosted for the fourth time.
However, an overview on capability- and skill based systems in factory automation and manufacturing systems is missing.
This paper aims to provide such an overview and give insights to this particular field of research.
We conducted a concise literature survey of papers covering the topics of capabilities and skills in manufacturing from the last ten years of the \gls{etfa} conference.
We found 247 papers with a notion on capabilities and skills and identified and analyzed 34 relevant papers which met this survey's inclusion criteria.
In this paper, we provide (i) an overview of the research field, (ii) an analysis of the characteristics of capabilities and skills, and (iii) a discussion on gaps and opportunities.
\end{abstract}


\glsresetall

\section{Introduction} 
\label{sec:Intro}


As customer requirements toward production tend to change more frequently, it becomes necessary to pursue flexible and variable production approaches.
Industry 4.0\footnote{Research agenda Industrie 4.0: \url{https://bit.ly/36mMbYu}} envisions \glspl{cpps} that facilitate an automated and adaptive production of mass-customizable products.
To achieve adaptability, \glspl{cpps} use modern production techniques and can interact with their environment using the latest IT~\cite{Monostori2014}.
\gls{cpps} engineering and operation requires to model the products, production processes, and production resources.
A popular modeling approach for this purpose is the \gls{ppr} concept~\cite{schleipen2015requirements}.

However, the selection of proper production resources for a given production process is a complex, challenging, and time-consuming task.
To foster enhanced adaptability and tackle these challenges, \emph{skill-based engineering} goes a step further in modeling.
It aims to abstract and decouple between the production processes and their requirements, e.g., production or quality criteria, and the resources that execute the processes.

The IEEE Conference \gls{etfa} is a well-established conference in the research field of factory automation and its links to computer science.
This year, for the fourth time, \gls{etfa} hosts a special session on capabilities \& skills in manufacturing.
While the special session already brings together academics and practitioners and condenses research, skill-based engineering was already brought up earlier in the \gls{etfa} community.

With this contribution, we aim to answer the following research questions:
\textbf{\emph{RQ1:}}  What is the current state of research of capabilities \& skills?, \textbf{\emph{RQ2:}} What are common understandings of capabilities \& skills in research?, and \textbf{\emph{RQ3:}} What are open gaps and opportunities for further research?

The goal of this paper is to provide a structured overview concerning skill-based manufacturing research from industry and academia.
Therefore and to identify recent topics and open issues for research, we conducted a concise literature survey with contributions of the last ten years of \gls{etfa}.
Out of 247 papers, we identified and analyzed 34 relevant papers. 
Our main contributions are: (i) a survey of 34 selected papers published at \gls{etfa}, (ii) a current state and overview of the research field, (iii) an analysis of the characteristics of the publications, especially, considering the definition of skills and capabilities, and (iv) a discussion on gaps and opportunities of the research field of skills and capabilities.

The remainder of the paper is structured as follows.
Section~\ref{sec:Methodology} describes methodology used before Section~\ref{sec:Analysis} presents the results of our analysis.
Section~\ref{sec:Analysis} investigates the survey results for key aspects in regards to skills and capabilities.
Sections~\ref{sec:Discussion} and \ref{sec:Conclusion} discusses the survey and draws conclusions.

\section{Methodology} 
\label{sec:Methodology}



\subsection{Survey Rationale and Protocol}

This year, the IEEE Conference \gls{etfa} will host the fourth special session on \emph{skill-based systems engineering}.
We took this opportunity as an incentive to conduct a concise literature survey that aims to provide an overview of publication on \emph{skills} and \emph{capabilities} at \gls{etfa} over the last ten years.
We argue, that \gls{etfa} with its focus on information technology, artificial intelligence and information modelling in the context of automation is an ideal foundation for such a survey.

To conduct the survey, we adapted a prominent \gls{slr} guideline by Kitchenham \emph{et. al}~\cite{Kitchenham2016} to our purpose.
We first defined a research protocol to ensure the quality of the survey.
We then defined research questions to be addressed by the literature survey, a search process, and inclusion and exclusion criteria to filter the found publications.

\subsection{Search String and Search Process}

Well-defined keywords and a search string are crucial for the quality of a literature survey.
All \gls{etfa} publications are accessible through \emph{IEEE Xplore}\footnote{IEEE Xplore: \url{https://ieeexplore.ieee.org}} and can be efficiently searched and filtered using a powerful search tool.
Therefore, we defined the keywords and search string confirming to their search syntax.
We defined the keywords based on an initial set of papers on \emph{skill-based systems engineering}.
Listing~\ref{listing:searchQuery} shows our search string, which can be split into two groups.

\begin{lstlisting}[language=SQL, caption=The \emph{search string} for the survey., label=listing:searchQuery]
(("All Metadata": Skill OR Capability) 
  AND 
(("Publication Title": Emerging Technologies AND Factory Automation)))
\end{lstlisting}

The keywords in the first groups aim to find publications with the terms \emph{skill} or \emph{capability} in all metadata, such as title and abstract.
\emph{IEEE Xplore} expands these keywords to their plural also searching for them as parts of a word.
The second part constrains the search to the publications of \gls{etfa} (which had its title written in different ways over the years).

The resulting publications undergo the following process to identify the relevant works. 
First, we filter for results that were contributed to ETFA within the last ten years. 
Afterwards, we go through the publications and group them into related, unclear, and unrelated publications by reading the titles and abstracts.
We discuss the unclear papers and, if necessary, skim through their text and decide which group we assign them to.
The related publications are read and analyzed in detail.

\subsection{Data Collection and Analysis}

We defined 16 data items to extract data from the publications.
Data items D1 to D5 were extracted as metadata for the particular papers, i.e., authors, title, year, abstract, and keywords.  
Data item D6 aims at extracting the used capability or skill \emph{definition}.
D7 collects the \emph{requirements} that should be fulfilled by capabilities \& skills.
D8 records the \emph{modelling concepts} used to represent capabilities \& skills.
D9 aims at extracting the \emph{solution} approach presented in the paper.
D10 collects the \emph{technologies} used to model capabilities \& skills or implement a solution.
D11 collects the \emph{use cases} used in the publications.
D12 records the \emph{type of evaluation}, e.g., conceptual or empirical, to assess the solution approach in a paper.
In D13, we estimate the \emph{technology readiness level}~\cite{Heder2017} of the proposed solutions to reason on their maturity.
D14 aims to extract the envisioned \emph{benefits} that should be brought to manufacturing.
D15 extracts remaining \emph{challenges} that need to be solved.
D16 identifies \emph{future work} in research.


We used collaborative spreadsheets to collate and store the data, making them available to the other authors.
The data were manually reviewed and analyzed in the spreadsheets using, e.g., frequency analysis.
Thereby we were able to gain insights and answer the research questions of the survey.







\subsection{Conducting the Survey}



\begin{figure}[ht]
    \footnotesize
    \centering
	\includegraphics[width=\columnwidth]{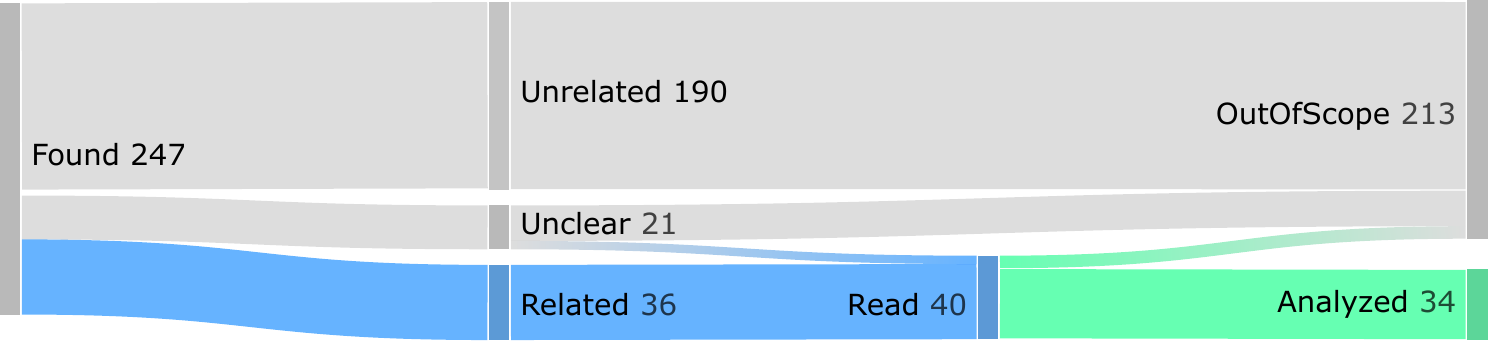}
    \caption{The selection flow resulting in 34 out of 247 publications.}
    \label{fig:literature-sankey}
\end{figure}

An overview of our actual search and selection process is given in Figure~\ref{fig:literature-sankey}.
In total, \gls{etfa}, which is held yearly since 1992, published over 5400 papers at its 26 conferences, with over 2700 paper in the last ten years.
The search string (cf. Listing~\ref{listing:searchQuery}) retrieved 247 publications from the last ten years.

Both \emph{skill} and \emph{capability} are mostly used in a general form to denote abilities of all kinds of individuals (e.g. innovation capabilities of companies or worker skills).
To filter out the use of the terms, we manually filtered the 247 results by reading the title and abstract analyzing their uses.


Through the filtering process, 190 papers were marked as \emph{unrelated} while 36 were marked as \emph{related}.
The remaining 21 \emph{unclear} papers underwent an additional filtering process, in which we skimmed the full texts and searched for occurrences of \emph{capability} or \emph{skill}.
From the unclear papers, 17 where found to be \emph{out of scope} and 4 were added to the set to \emph{read}.
During detailed reading and data collection, we found that 6 publications still did not fit to the scope of this study, resulting in a total of 34 \emph{analyzed} papers that can also be found online.\footnote{Additional online material: \url{https://github.com/tuw-qse/etfa-skills-survey}}


\section{Analysis of Skill and Capability Research} 
\label{sec:Analysis}
The following sections describe various aspects of the selected papers aiming to answer the research questions.

\subsection{Meta Analysis of the Publications}
\label{sec:meta-analysis}
\begin{figure}[ht]
    \centering
	\includegraphics[width=.45\textwidth]{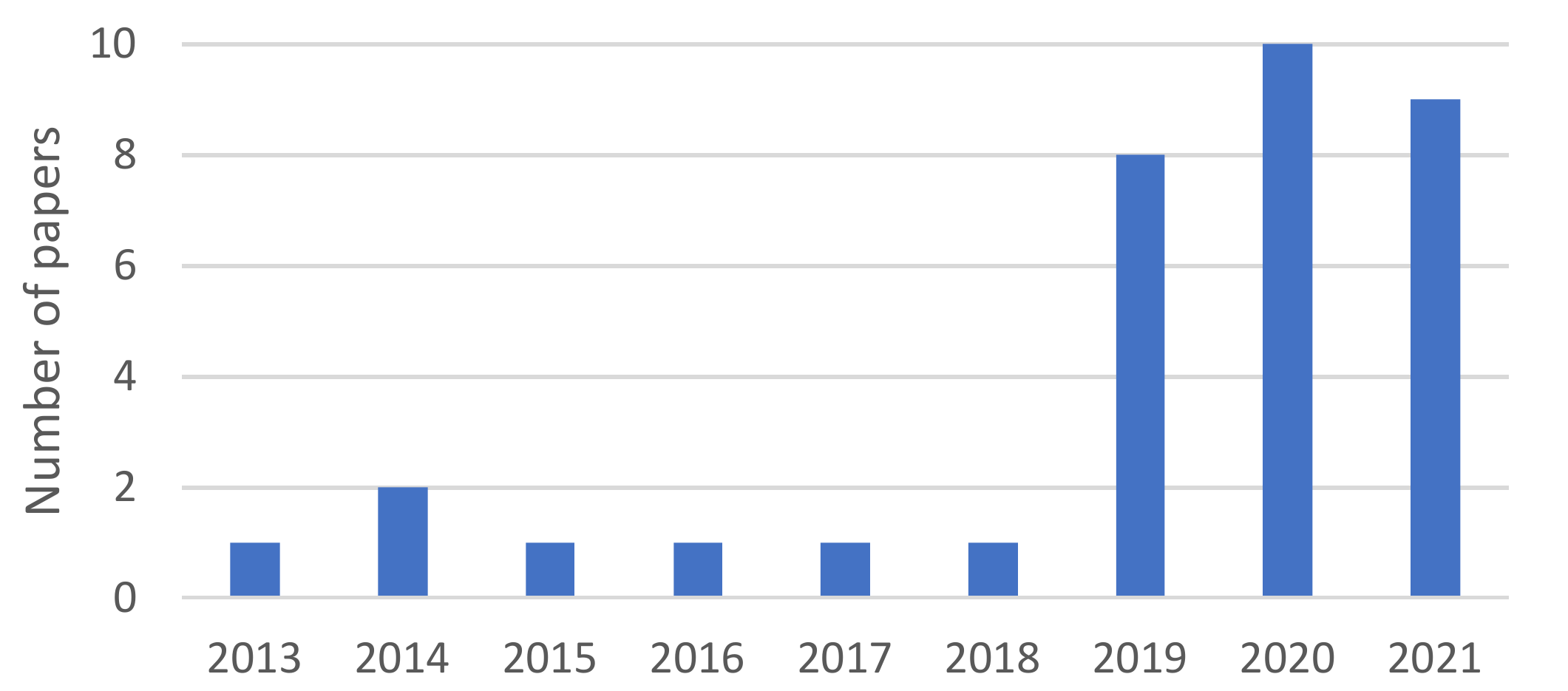}
    \caption{Distribution of selected publications over the last decade.}
    \label{fig:papers_per_year}
\end{figure}

\begin{figure}[ht]
    \centering
    \includegraphics[width=.30\textwidth]{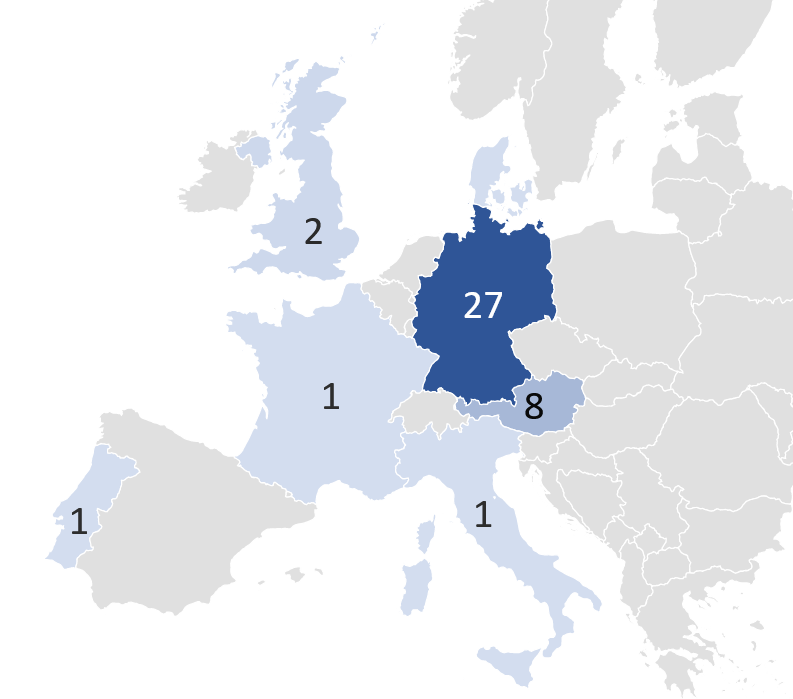}
    \caption{Distribution of selected publications per country.}
    \label{fig:papers_per_country}
\end{figure}

\begin{figure}[ht]
    \centering
    \includegraphics[width=\linewidth]{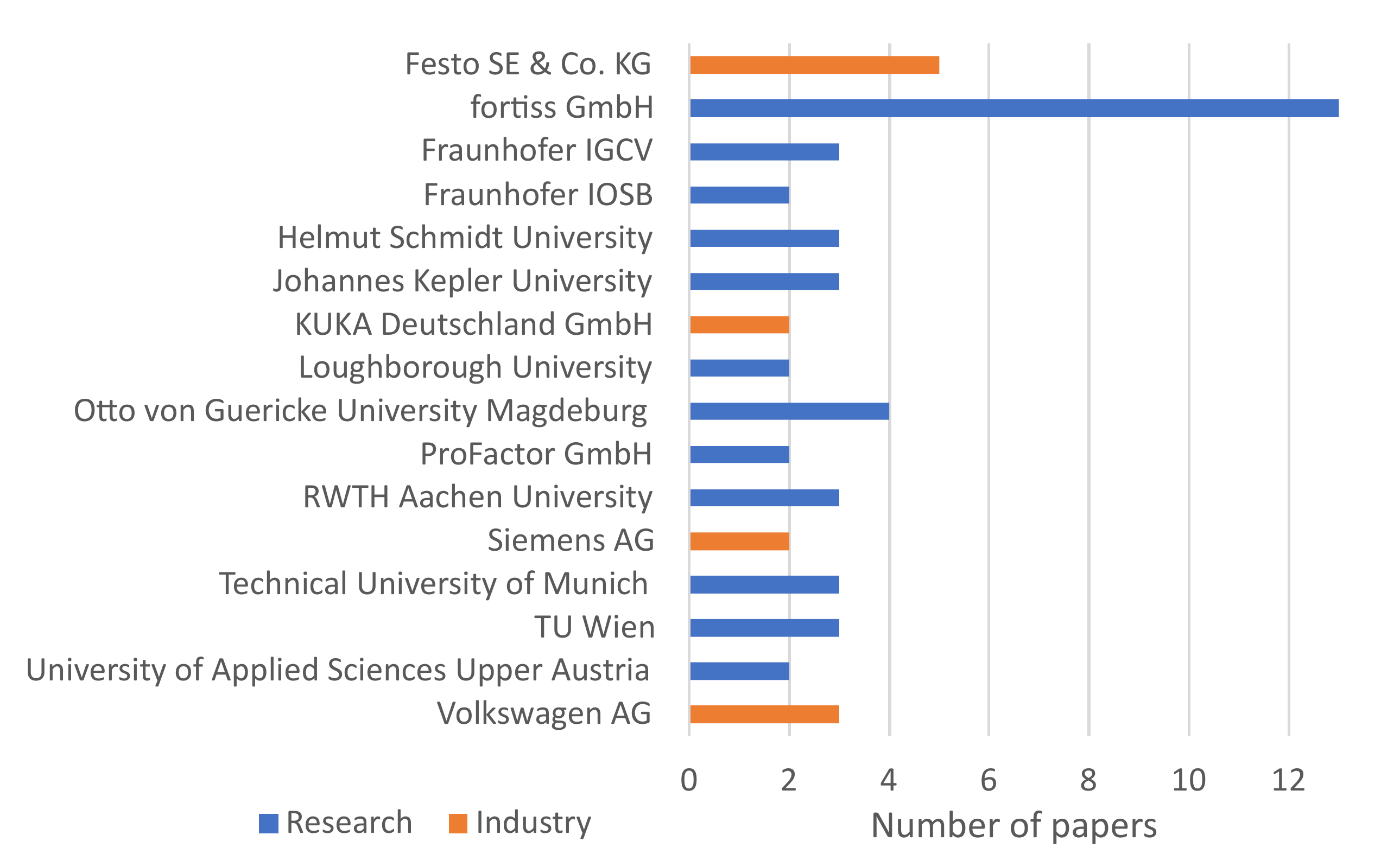}
    \caption{Breakdown of institutions with more than one publication}
    \label{fig:papers_per_institution}
\end{figure}

Looking at the number of papers per year, there is a clear accumulation in the last three years (cf. Figure~\ref{fig:papers_per_year}), although some papers covered the topic of skills and capabilities before.
Figure~\ref{fig:papers_per_country} reveals that the individual institutions publishing the papers are all located in Europe, especially, in the German-speaking regions. The proportion of German and Austrian authors in this research field is significantly higher than their proportion on all ETFA publications of the last ten years ($\sim$33\%).
Figure~\ref{fig:papers_per_institution} shows the distribution of all institutions involved.
It should be noted that the proportion of industrial companies is significantly high, which suggests that interest in and use cases from the industry are definitely present.



\subsection{Motivation for Capabilities \& Skills} 
\label{sec:Motivation}

Most publications explicitly discuss motivations for research on capabilities \& skills in manufacturing.
These motivations fall mainly into four categories.
The motivation ranking first, mentioned by 25 papers, is the \emph{need for adaptability or flexibility} in manufacturing, e.g., \cite{Pfrommer2013_PPRSProductionskills,Pfrommer2014_Modellingorchestrationservice,Sonnleithner2021_IEC61499Distributed,Meixner2020_ModelingExpertKnowledge}, underpinning the motivation in Section~\ref{sec:Intro}.
The challenge of \emph{product variability} was raised in 23 publications~\cite{Herzog2020_AllocationPPRSplant, Hoang2019_CapabilityModelAdaptation, Motsch2021_ConceptModelingUsage}.
Additionally, twelve publications noted that the decreasing product batch size, a.k.a., \emph{lot size 1}, is challenging~\cite{Koecher2020_FormalCapabilitySkill,Zimmermann2021_ConceptSelectingSuitable,Spitzer2020_genericApproachIndustrial}.
These two issues, leading to frequently changing requirements toward production \cite{Christian2019_Dynamicintegrationmanual}, might be tackled by adaptability.
The fourth challenge for \gls{cpps} engineering mentioned by nine papers, e.g., \cite{Herzog2020_AllocationPPRSplant,Weser2020_OntologybasedMetamodel,Zimmermann2021_ConceptSelectingSuitable}, is a required increase of planning efficiency.
While most publications consider these concepts as motivations, some already see them as expected benefits (cf. Section~\ref{sec:Benefits}).

\subsection{Requirements toward Capabilities \& Skills} 
\label{sec:Requirements}
Capabilities \& skills should foster the Industry~4.0 vision of \gls{cpps} flexibility.
To design suitable approaches, academia must address particular \emph{requirements} towards skills and capabilities.
In 21 of 34 publications, requirements are mentioned either implicitly or explicitly.
In Table~\ref{tab:requirements} we condensed the mentioned requirements and the corresponding publications.

{\renewcommand{\arraystretch}{1.1}
\begin{table}[ht]
\centering
\caption{Requirements towards skill-based systems engineering.}
\begin{tabularx}{\columnwidth}{Xl}
Requirement                       & Publication(s) \vspace{0.1em} \\ \hline 
Formal Description                & \hspace{1sp}\citeS{Keddis2014_Capabilitybasedplanning,Malakuti2018_ChallengesSkillbased,Profanter2019_HardwareAgnosticOPC,Dorofeev2019_EvaluatingSkillBased,Koecher2020_AutomatingDevelopmentMachine,Himmelhuber2020_OntologyBasedSkill,Eymueller2021_TowardsRealTime} \\
(Re-)Configurability/Adaptability & \hspace{1sp}\citeS{Dorofeev2017_Deviceadapterconcept,Profanter2019_HardwareAgnosticOPC,Dorofeev2019_AgileOperationalBehavior,Mayrhofer2020_CapabilityBasedProcess,Huang2021_AASCapabilityBased} \\
Matchability                      & \hspace{1sp}\citeS{Malakuti2018_ChallengesSkillbased,Herzog2020_AllocationPPRSplant,Meixner2020_ModelingExpertKnowledge,Sarna2021_ReducingRiskIndustrial} \\
Statefulness/Deterministic        & \hspace{1sp}\citeS{Spitzer2021_ATLASGenericFramework,Profanter2019_HardwareAgnosticOPC,Eymueller2021_TowardsRealTime} \\
Classifiability                   & \hspace{1sp}\citeS{Malakuti2018_ChallengesSkillbased,Profanter2019_HardwareAgnosticOPC} \\
Communication Interface           & \hspace{1sp}\citeS{Zimmermann2019_SkillbasedEngineering,Eymueller2021_TowardsRealTime} \\
Modularity                        & \hspace{1sp}\citeS{Dorofeev2019_EvaluatingSkillBased,Soerensen2020_TowardsDigitalTwins} \\
Vendor-Neutrality                 & \hspace{1sp}\citeS{Koecher2020_FormalCapabilitySkill,Eymueller2021_TowardsRealTime} \\
Identifyability                   & \hspace{1sp}\citeS{Koecher2020_FormalCapabilitySkill,Profanter2019_HardwareAgnosticOPC} \\
Executability                     & \hspace{1sp}\citeS{Profanter2019_HardwareAgnosticOPC,Dorofeev2017_Deviceadapterconcept} \\
(Auto-)Discovery                  & \hspace{1sp}\citeS{Dorofeev2017_Deviceadapterconcept} \\
Reusability                       & \hspace{1sp}\citeS{Dorofeev2019_EvaluatingSkillBased} \\
Scalability                       & \hspace{1sp}\citeS{Dorofeev2019_AgileOperationalBehavior} \\
Extensibility                     & \hspace{1sp}\citeS{Koecher2020_FormalCapabilitySkill}           
\end{tabularx}
\label{tab:requirements}
\end{table}
}

The primary requirement, raised in seven publications, was the need for a \emph{formal description} of capabilities \& skills, e.g., in a machine-readable format.
Second, capabilities \& skills should be prepared to facilitate the \emph{(re-)configurability} of the resources and the \gls{cpps}.
Another important requirement is the ability to (automatically) \emph{match} the production process requirements with the functions provided by resources.
\emph{Statefulness} or the \emph{deterministic} progression of a skill is a requirement mentioned in three publications.

\subsection{Definitions of Capabilities \& Skills} 
\label{sec:Definition}

To the best of our knowledge and following the discussions of previous \gls{etfa} special sessions, there is no consent on the definition of the terms \emph{capability} \& \emph{skill}. 
This section presents the collected and categorized characteristics for capability \& skill definitions from the publications.

{\renewcommand{\arraystretch}{1.1}
\begin{table*}[ht]
\caption{Classification of Skill and Capability Definition Characteristics. \textsuperscript{+} \dots varying \emph{skill} definition, \textsuperscript{*} \dots distinction of \emph{skills} and \emph{capabilities}, \textsuperscript{a} \dots called provider and consumer, \textsuperscript{b} \dots implicitly mentioned, \textsuperscript{c} \dots \emph{skills} are composed to actions (and tasks), \textsuperscript{d} \dots uses \emph{capability} as concept, \textsuperscript{e} \dots distinguished via \emph{skills} and \emph{capabilities}, \textsuperscript{g} \dots combined via \emph{skills} and \emph{actions}} 
\begin{tabularx}{\textwidth}{ll|ll|ccl|ll}
Year     & Publication                                            & \multicolumn{2}{c|}{Distinction\textsuperscript{*}}                                   & \multicolumn{3}{c|}{Definition}             & Atomic                                       & Required                 \\
 &                                          & Without                                & With & Without & With & With external reference & Composite                       & Provided                                    \\ \hline
2013 & Pfrommer \emph{et. al}~\citeS{Pfrommer2013_PPRSProductionskills}         & ~$\times$         &                   &               & ~$\times$          &               & ~$\times$                                      &                  \\
2014 & Keddis \emph{et. al}~\citeS{Keddis2014_Capabilitybasedplanning}         & ($\times$)\textsuperscript{d}         &                   &               & ~$\times$          &               & ~$\times$                                      &                  \\
 & Pfrommer \emph{et. al}~\citeS{Pfrommer2014_Modellingorchestrationservice} & ~$\times$         &                   &               &             & Pfrommer \emph{et. al}~\citeS{Pfrommer2013_PPRSProductionskills}               &                                        &                  \\
2015 & Keddis \emph{et. al}~\citeS{Keddis2015_Optimizingschedulesadaptable}    & ($\times$)\textsuperscript{d}         &                   &               &            & Keddis \emph{et. al}~\citeS{Keddis2014_Capabilitybasedplanning}               &                                        &                  \\
2016 & Wenger \emph{et. al}~\citeS{Wenger2016_modelbasedengineering}\textsuperscript{+}         & ~$\times$         &                   &               & ~$\times$          &               & ~$\times$          &                  \\
2017 & Dorofeev \emph{et. al}~\citeS{Dorofeev2017_Deviceadapterconcept}          & ~$\times$         &                   &               &             & Pfrommer \emph{et. al}~\citeS{Pfrommer2013_PPRSProductionskills}               & ~$\times$                                      &                  \\
2018 & Malakuti \emph{et. al}~\citeS{Malakuti2018_ChallengesSkillbased}          & ~$\times$         &                   &               &             & Industrie 4.0 – Glossary~\citeS{PlattformIndustrie40a}               & ~$\times$                                      & ~$\times$                \\
2019 & Christian \emph{et. al}~\citeS{Christian2019_Dynamicintegrationmanual}     & ~$\times$         &                   & ~$\times$         &           &                & ~$\times$         &                  \\
 & Dorofeev \emph{et. al}~\citeS{Dorofeev2019_AgileOperationalBehavior}      & ~$\times$         &                   &               &                                        & Pfrommer \emph{et. al}~\citeS{Pfrommer2014_Modellingorchestrationservice}               & ~$\times$                                      &                  \\
 & Dorofeev \& Wenger~\citeS{Dorofeev2019_EvaluatingSkillBased}          & ~$\times$         &                   &               &                         & Pfrommer \emph{et. al}~\cite{Pfrommer2015}               &                                        & ($\times$)\textsuperscript{a}   \\
 & Evers \emph{et. al}~\citeS{Evers2019_RoadmapSkillBased}                & ~$\times$         &                   &               & ~$\times$          &               &               & ~$\times$              \\
 & Hoang \& Fay~\citeS{Hoang2019_CapabilityModelAdaptation}        & ($\times$)\textsuperscript{d} &                   &               &             & Malakuti \emph{et. al}~\citeS{Malakuti2018_ChallengesSkillbased}, Industrie 4.0 – Glossary~\citeS{PlattformIndustrie40a}               &                                        &                  \\
 & Kathrein \emph{et. al}~\citeS{Kathrein2019_EfficientProductionSystem}     & ~$\times$         &                   &               &             & Pfrommer \emph{et. al}~\citeS{Pfrommer2013_PPRSProductionskills}               &                                        & ~$\times$                \\
 & Profanter \emph{et. al}~\citeS{Profanter2019_HardwareAgnosticOPC}          & ~$\times$         &                   &               &             & Dorofeev \emph{et. al}~\citeS{Dorofeev2017_Deviceadapterconcept}               & ~$\times$                                      &                  \\
 & Zimmermann \emph{et. al}~\citeS{Zimmermann2019_SkillbasedEngineering}       & ~$\times$         &                   &               &             & Industrie 4.0 – Glossary~\citeS{PlattformIndustrie40a}               & ~$\times$                                      & ($\times$)\textsuperscript{a}   \\
2020 & Herzog \emph{et. al}~\citeS{Herzog2020_AllocationPPRSplant}             & ~$\times$         &                   &               &             & Pfrommer \emph{et. al}~\citeS{Pfrommer2013_PPRSProductionskills}               & ~$\times$                                      & ~$\times$                \\
 & Heuss \& Reinhart~\citeS{Heuss2020_IntegrationAutonomousTask}\textsuperscript{+}      & ~$\times$         &                   &               & ~$\times$          &               & ~$\times$                                       &                  \\
 & Himmelhuber \emph{et. al}~\citeS{Himmelhuber2020_OntologyBasedSkill}         & ~$\times$         &                   &               &                                   & Hoang \emph{et. al}~\cite{Hoang2018}               &                                        & ~$\times$                \\
 & Koecher \emph{et. al}~\citeS{Koecher2020_AutomatingDevelopmentMachine}   &           & ~$\times$                 &               &                                 & Perzylo \emph{et. al}~\cite{Perzylo2019}               &                                        & ($\times$)\textsuperscript{e} \\
 & Koecher \emph{et. al}~\citeS{Koecher2020_FormalCapabilitySkill}          &           & ~$\times$                 &               & ~$\times$          &               &                                        & ($\times$)\textsuperscript{e} \\
 & Mayrhofer \emph{et. al}~\citeS{Mayrhofer2020_CapabilityBasedProcess}     & ($\times$)\textsuperscript{d}         &                   &               & ~$\times$          &               &                                        &                  \\
 & Meixner \emph{et. al}~\citeS{Meixner2020_ModelingExpertKnowledge}        & ~$\times$         &                   &               &             & Pfrommer \emph{et. al}~\citeS{Pfrommer2013_PPRSProductionskills}               & ~$\times$                                      & ~$\times$                \\
 & Soerensen \emph{et. al}~\citeS{Soerensen2020_TowardsDigitalTwins}\textsuperscript{+}        & ~$\times$         &                   &               & ~$\times$          &               &                                        &                  \\
 & Spitzer \emph{et. al}~\citeS{Spitzer2020_genericApproachIndustrial}      & ~$\times$         &                   &               &             & Malakuti \emph{et. al}~\citeS{Malakuti2018_ChallengesSkillbased}               & ~$\times$                                      & ($\times$)\textsuperscript{g}  \\
 & Weser \emph{et. al}~\citeS{Weser2020_OntologybasedMetamodel}           &           & ~$\times$                 &               & ~$\times$          &               & ~$\times$                                      & ($\times$)\textsuperscript{e} \\
2021 & Dorofeev \emph{et. al}~\citeS{Dorofeev2021_GenerationOrchestratorCode}    & ~$\times$         &                   &               &                                   & Bayha \emph{et. al}~\cite{Bayha2020}               & ~$\times$                                      & ~$\times$                \\
 & Eymueller \emph{et. al}~\citeS{Eymueller2021_TowardsRealTime}              &           & ~$\times$                 &               & ~$\times$          &               & ~$\times$                                      & ($\times$)\textsuperscript{e} \\
 & Huang \emph{et. al}~\citeS{Huang2021_AASCapabilityBased}               &           & ~$\times$                 & ~$\times$             &           &                & ~$\times$                                      & ($\times$)\textsuperscript{e} \\
 & Motsch \emph{et. al}~\citeS{Motsch2021_ConceptModelingUsage}            &           & ~$\times$                 &               &                                   & Bayha \emph{et. al}~\cite{Bayha2020}, Pfrommer \emph{et. al}~\citeS{Pfrommer2013_PPRSProductionskills}               & ($\times$)\textsuperscript{b}         & ($\times$)\textsuperscript{e} \\
 & Sarna \emph{et. al}~\citeS{Sarna2021_ReducingRiskIndustrial}           & ~$\times$         &                   &               &           & Meixner \emph{et. al}~\citeS{Meixner2020_ModelingExpertKnowledge}               &                                        & ~$\times$                \\
 & Sonnleithner \emph{et. al}~\citeS{Sonnleithner2021_IEC61499Distributed}       &           &                   &               &  & Koecher \emph{et. al}~\citeS{Koecher2020_AutomatingDevelopmentMachine}, Spitzer \emph{et. al}~\citeS{Spitzer2020_genericApproachIndustrial}               & ~$\times$                                      &                  \\
 & Spitzer \emph{et. al}~\citeS{Spitzer2021_ATLASGenericFramework}          & ~$\times$         &                   &               & ~$\times$          &               &                                        &                  \\
 & Villagrossi \emph{et. al}~\citeS{Villagrossi2021_Simplifyrobotprogramming}\textsuperscript{+} & ~$\times$         &                   & ~$\times$              &           &                & ($\times$)\textsuperscript{c} &                  \\
 & Zimmermann \emph{et. al}~\citeS{Zimmermann2021_ConceptSelectingSuitable}    &           & ~$\times$                 &               & ~$\times$          &               & ~$\times$                                      & ($\times$)\textsuperscript{e}
\end{tabularx}
\label{tab:definitions}
\end{table*}
}

Table~\ref{tab:definitions} presents the classification of capability \& skill definitions from the selected publication sorted by year and first author.
The publications marked with a \textsuperscript{+} have a definition varying from the other definitions.
For instance, Wenger \emph{et. al}~\cite{Wenger2016_modelbasedengineering} use a graph-based definition based, i.e., nodes and their relations, to represent a particular functionality.

The column \emph{distinction} indicates whether the definition in the publication distinguishes between capabilities \& skills.
Publications marked with \textsuperscript{d} use the term \emph{capability} rather than \emph{skill}.
The data shows that a clear distinction was only made by seven publications in the more recent years.

Column \emph{definition} specifies if a publication provides no clear definition, its own definition, or uses a definition from related work.
Only three publications do not provide a definition, 12 papers provide their own definition or a definition that cannot be traced back to related work and 19 papers refer to definitions from related work.
The most cited definition is the one by Pfrommer \emph{et. al.}~\cite{Pfrommer2013_PPRSProductionskills}, followed by the one of the Industrie~4.0 Glossary resp. by Malakuti \emph{et. al}~\cite{Malakuti2018_ChallengesSkillbased}.

The column \emph{atomic/composite} shows if the publication uses the concept of atomic capabilities \& skills, e.g., gripping, that can be combined to composite ones, e.g., pick and place.
While Motsch \emph{et. al}~\cite{Motsch2021_ConceptModelingUsage} mention this ability implicitly, Villagrossi \emph{et. al}~\cite{Villagrossi2021_Simplifyrobotprogramming} use the terms actions and tasks.
The data show that composition is a common theme in the publications analyzed.

Column \emph{required/provided} shows if a publication distinguishes between capabilities \& skills that are required, e.g., by processes, and provided, e.g., by resources.
Some publications use synonyms, such as provider and consumer (labeled with~\textsuperscript{a}) or skills and actions (labeled with~\textsuperscript{g}).
Several publications clearly distinguish between capabilities \& skills (labeled with~\textsuperscript{e}).
The data indicate that the required and provided skill concept is widely used in early work already but is superseded by the concept of separated capabilities \& skills.
One reason seems to be that capabilities are considered an abstract form of describing requirements.

{\renewcommand{\arraystretch}{1.1}
\begin{table*}[ht]
\caption{Definition of skills or capabilities in literature. \textsuperscript{+} \dots origin of definition}
\begin{tabularx}{\textwidth}{lXp{2.5cm}}
Key Point                                          & Definition                                                                                                                                                                                                            & Publication(s) \\ \hline
Perform a process                                  & -- A skill is the ability of a resource to perform a process and thus the relation of process and resource, enriched with additional information.                                                                        & \citeS{Pfrommer2013_PPRSProductionskills}\textsuperscript{+}, \citeS{Sarna2021_ReducingRiskIndustrial,Motsch2021_ConceptModelingUsage,Meixner2020_ModelingExpertKnowledge,Herzog2020_AllocationPPRSplant,Profanter2019_HardwareAgnosticOPC,Kathrein2019_EfficientProductionSystem,Dorofeev2019_EvaluatingSkillBased,Dorofeev2019_AgileOperationalBehavior,Dorofeev2017_Deviceadapterconcept,Pfrommer2014_Modellingorchestrationservice}           \\ \hline
\multirow{2}{*}{Achieve an effect}                 & -- A skill is defined as the potential of a production resource to achieve an effect within a domain.                                                                                                                   & \cite{PlattformIndustrie40a}\textsuperscript{+}, \citeS{Sonnleithner2021_IEC61499Distributed,Spitzer2020_genericApproachIndustrial,Himmelhuber2020_OntologyBasedSkill,Zimmermann2019_SkillbasedEngineering,Hoang2019_CapabilityModelAdaptation,Malakuti2018_ChallengesSkillbased}  \\
                                                   & -- An abstract capability describes the ability of an entity to perform a specific activity.                                                                                                                               & \citeS{Mayrhofer2020_CapabilityBasedProcess}\textsuperscript{+}           \\ \hline
\multirow{6}{*}{Capability/Skill} & -- A capability is an abstract description of a process provided by a machine. A skill is an executable function that might be used to execute such a process on a machine.                          & \citeS{Koecher2020_AutomatingDevelopmentMachine}\textsuperscript{+}, \citeS{Sonnleithner2021_IEC61499Distributed}           \\
                                                   & -- A capability is defined as a machine’s potential to execute some kind of transformation or process. In this context, skills are compared to skill requirements and selected to create a new or reconfigured system.   &    \citeS{Koecher2020_FormalCapabilitySkill}\textsuperscript{+}        \\
                                                   & -- A skill is defined as an executable implementation of a capability offered by a device or service where capabilities act as description of the functionality of a component.                                          &    \citeS{Weser2020_OntologybasedMetamodel}\textsuperscript{+}        \\
                                                   & -- Skills are often referred to as abstracting the functionality of the control devices by using generic interfaces. Skills should offer an interface to communicate with a wide range of heterogeneous control devices. &    \citeS{Dorofeev2021_GenerationOrchestratorCode}\textsuperscript{+}, \citeS{Motsch2021_ConceptModelingUsage}        \\
                                                   & -- A skill is a executable capability of one or more resources.                                                                                                                                                          &    \citeS{Eymueller2021_TowardsRealTime}\textsuperscript{+}        \\
                                                   & -- Skills are implemented as kind of service provided by specific hardware throughout the production plant.                                                                                         &    \citeS{Spitzer2021_ATLASGenericFramework}\textsuperscript{+}        \\ \hline
\multirow{4}{*}{Resource-based}                    & -- Resources provide and implement capabilities that are required for the production. The required capabilities are described in production plans.                                                                       &    \citeS{Keddis2014_Capabilitybasedplanning}\textsuperscript{+}, \citeS{Keddis2015_Optimizingschedulesadaptable}         \\
                                                   & -- Interpreting and realizing each process as some functionalities a component/product can perform is called a skill of the components.                                                                                  &    \citeS{Evers2019_RoadmapSkillBased}\textsuperscript{+}       \\
                                                   & -- Skill refers to the functionalities that a production machine provides. These skill descriptions are the basis for the control functionality of the production process.                                                &    \citeS{Himmelhuber2020_OntologyBasedSkill}\textsuperscript{+}        \\
                                                   & -- Skills or capabilities are used to describe the functionality of resources and simplify the matching process                                                                                                          &    \citeS{Zimmermann2021_ConceptSelectingSuitable}\textsuperscript{+}        \\ \hline
\multirow{4}{*}{Other}                             & -- A skill is a composition of nodes and their relations where a node specifies interfaces to exchange typed messages according to the publish/subscribe paradigm.                                                       &    \citeS{Wenger2016_modelbasedengineering}\textsuperscript{+}        \\
                                                   & -- Functionalities or services offered by a system, which are also referred to as the skills of components.                                                                                                  &    \citeS{Heuss2020_IntegrationAutonomousTask}\textsuperscript{+}        \\
                                                   & -- Skills are used as abstract programming tools for visual robot programming.                                                                                                                                            &    \citeS{Soerensen2020_TowardsDigitalTwins}\textsuperscript{+}        \\
                                                   & -- Skill is the elementary movement executed by the robotic system.                                                                                                                                                      &    \citeS{Villagrossi2021_Simplifyrobotprogramming}\textsuperscript{+}       
\end{tabularx}
\label{tab:defs}
\end{table*}
}

From the classification of the definitions, we categorized the most prominent ones into groups.
We identified five groups of capability \& skill definitions with 17 different definitions.
Table~\ref{tab:defs} shows these groups with a key point, the particular definitions, and the publications that use the definition, including its source (marked with \textsuperscript{+}).
The groups concern the abilities to \emph{perform a process} or \emph{achieve an effect}, definitions that distinguish between \emph{capabilities \& skills}, those that use a \emph{resource-based} viewpoint, and \emph{other} definitions.


\subsection{Expected Benefits of Capabilities \& Skills} 
\label{sec:Benefits}
\begin{figure}[ht]
    \centering
	\includegraphics[width=.425\textwidth]{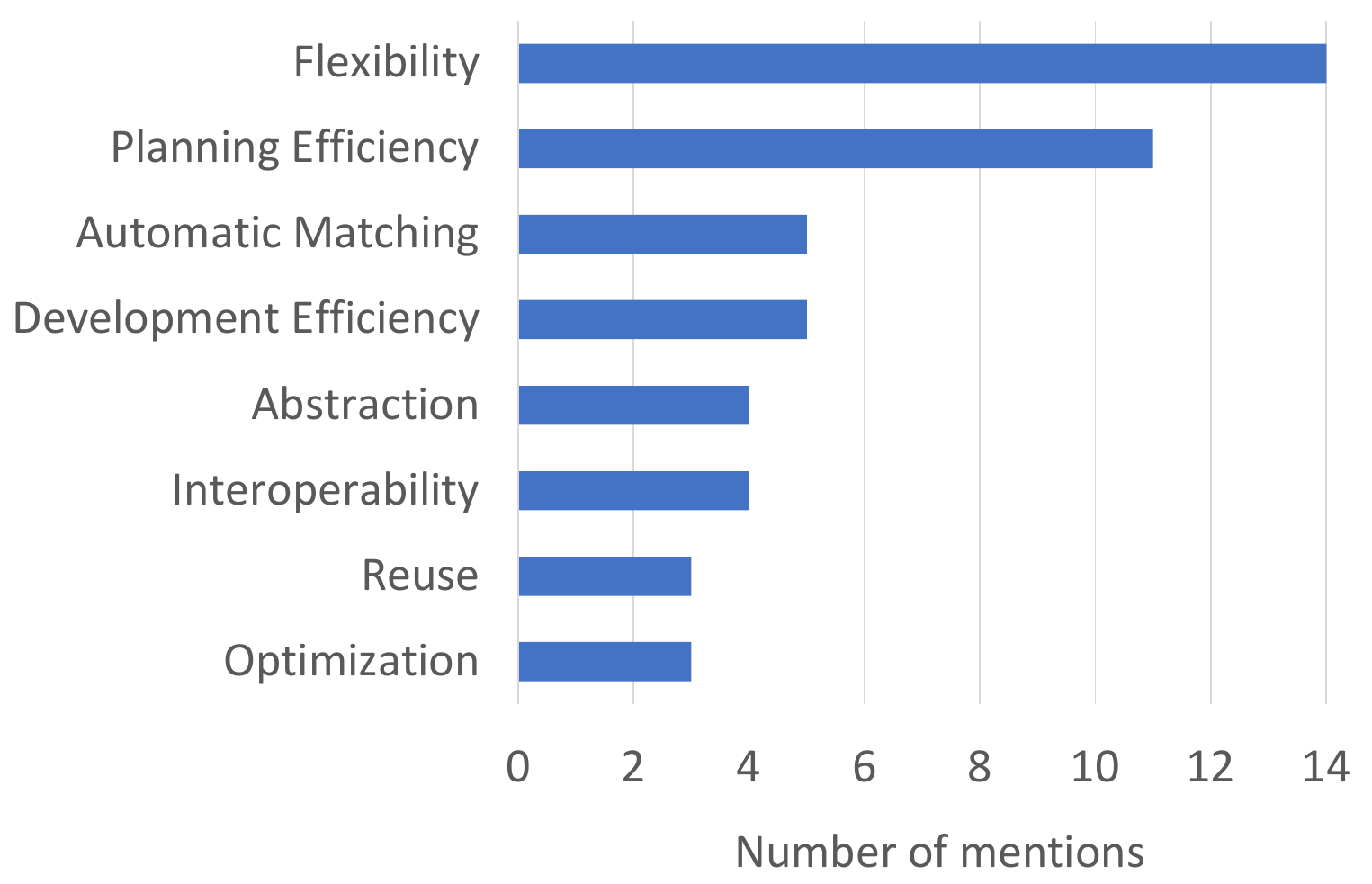}
    \caption{Overview of expected benefits.}
    \label{fig:benefits}
\end{figure}

Figure~\ref{fig:benefits} shows an overview of the eight most important benefits and how often they are mentioned in the papers.
The primary benefit \emph{flexibility} in manufacturing has been raised in 14 publications (e.g. \cite{Keddis2014_Capabilitybasedplanning, Malakuti2018_ChallengesSkillbased, Eymueller2021_TowardsRealTime}), targeting the requirements (re-)configurability and adaptability.
Second most, \emph{planning efficiency} in production planning has been mentioned (e.g. \cite{Himmelhuber2020_OntologyBasedSkill, Dorofeev2017_Deviceadapterconcept, Herzog2020_AllocationPPRSplant}). 
A subclass of \emph{planning efficiency} is \emph{automatic matching} (e.g. \cite{Meixner2020_ModelingExpertKnowledge,Hoang2019_CapabilityModelAdaptation,Weser2020_OntologybasedMetamodel}). 
Both benefits are addressing the requirements matchability, (re-)configurability and modularity. 
Five papers expose the benefit \emph{development efficiency}, expecting lower engineering costs and risks (\cite{Sarna2021_ReducingRiskIndustrial,Evers2019_RoadmapSkillBased}) and ease robot programs development (e.g. \cite{Dorofeev2021_GenerationOrchestratorCode, Villagrossi2021_Simplifyrobotprogramming}). 
The functional \emph{abstraction}, mentioned e.g. in \cite{Spitzer2021_ATLASGenericFramework,Soerensen2020_TowardsDigitalTwins,Pfrommer2013_PPRSProductionskills}, allows the separation of the product recipe/plan from the manufacturing execution.
This separation of concerns is the enabler for manufacturing \emph{flexibility} as the product recipes can be described technology-agnostic and their realization can be checked semantically through \emph{automatic matching}.
Also, functional \emph{abstraction} is the basis for two further benefits, \emph{interoperability} and \emph{reuse}.
\emph{Interoperability} is given as a manufacturing resources can be easily replaced by others with the same functionality (e.g. \cite{Profanter2019_HardwareAgnosticOPC,Christian2019_Dynamicintegrationmanual}).
This aspect is addressing especially the requirements regarding vendor-neutrality.
Product recipes and plans can be reused as they are independent of the resource execution (e.g. \cite{Dorofeev2019_EvaluatingSkillBased,Mayrhofer2020_CapabilityBasedProcess}).
The last benefit of capabilities \& skills in manufacturing is production \emph{optimization} (e.g. \cite{Herzog2020_AllocationPPRSplant,Zimmermann2021_ConceptSelectingSuitable}). Production plans can be optimized according to specific criteria, e.g., by a simulation (cf. \cite{Soerensen2020_TowardsDigitalTwins}).

\subsection{Use Cases, Application Domains and TRL} 
\label{sec:Usecase}
We also investigated how publications evaluated their approaches by analyzing both the considered use cases/processes and the evaluation. 
The use cases indicate that capabilities \& skills are researched almost exclusively in discrete manufacturing.
With \cite{Christian2019_Dynamicintegrationmanual}, concerning stem cell manufacturing, there is only one work that applies skills to process manufacturing.
For discrete manufacturing, there a broad range of manufacturing processes is considered for evaluation. 
Particularly noteworthy is that 11 of the 34 analyzed contributions are evaluated on simulated or real lab-scale plants with several capabilities forming a complete manufacturing process of a product, e.g., \cite{Sonnleithner2021_IEC61499Distributed, Dorofeev2019_AgileOperationalBehavior, Keddis2014_Capabilitybasedplanning}.
Six contributions apply their approaches to Pick \& Place, e.g., \cite{Spitzer2020_genericApproachIndustrial, Koecher2020_AutomatingDevelopmentMachine}. 
Five approaches consider assembly processes, e.g., \cite{Eymueller2021_TowardsRealTime, Meixner2020_ModelingExpertKnowledge}.
Five approaches do not consider specific processes at all. These are, e.g., contributions that highlight challenges \cite{Malakuti2018_ChallengesSkillbased} or include basic descriptions of meta-models \cite{Weser2020_OntologybasedMetamodel}. The remaining seven approaches use, e.g., commissioning, testing, transporting or soldering.


Looking at the evaluations, we found that 21 of the 34 publications implement their approaches as prototypes applying them to real or simulated machines.
Three approaches present a pure software implementation with a runtime analyses, e.g., \cite{Weser2020_OntologybasedMetamodel}.
Seven publications contain approaches on a conceptual level and discuss their application without tool support. 

Only five of the 34 publications (\cite{Koecher2020_FormalCapabilitySkill,Profanter2019_HardwareAgnosticOPC,Dorofeev2019_AgileOperationalBehavior, Koecher2020_AutomatingDevelopmentMachine,Villagrossi2021_Simplifyrobotprogramming}) publish models, algorithms, or data for their evaluation.
Thus, reproducibility and reusability of many approaches is only possible to a rather limited extent.

\subsection{Modeling Concepts (Solutions) and Technologies} 
\label{sec:Concepts}
Research on capabilities \& skills is often based on domain-specific models or modeling methods instead of generic modeling approaches such as UML. 
Domain-specific in the sense of the relevant papers means that production processes at different levels of automation starting from the field area up to generic product flow models are considered for modeling. 
 
There also exist several standards from ISA, VDI or VDMA that specify certain modeling elements.

In summary, the standardized modeling frameworks listed in Table~\ref{tab:technologies} are preferred and are the focus of this work.
{
    \renewcommand{\arraystretch}{1.1}
    \begin{table}[ht]
        \centering
        \caption{Technologies used for skill-based systems engineering.}
        \begin{tabularx}{\columnwidth}{Xl}
            Technology                   & Publication(s) \\
            \hline
            OPC~UA Information Model     & \hspace{1sp}\citeS{Spitzer2021_ATLASGenericFramework, Profanter2019_HardwareAgnosticOPC, Christian2019_Dynamicintegrationmanual, Eymueller2021_TowardsRealTime, Dorofeev2019_EvaluatingSkillBased} \\
            NAMUR MTP                    & \hspace{1sp}\citeS{Dorofeev2021_GenerationOrchestratorCode} \\
            AutomationML                 & \hspace{1sp}\citeS{Hoang2019_CapabilityModelAdaptation, Dorofeev2017_Deviceadapterconcept, Pfrommer2013_PPRSProductionskills, Dorofeev2019_AgileOperationalBehavior} \\
            Ontologies (e.g., \gls{owl}) & \hspace{1sp}\citeS{Koecher2020_FormalCapabilitySkill, Weser2020_OntologybasedMetamodel, Keddis2014_Capabilitybasedplanning, Malakuti2018_ChallengesSkillbased, Pfrommer2014_Modellingorchestrationservice, Himmelhuber2020_OntologyBasedSkill} \\
            \gls{emf}                    & \hspace{1sp}\citeS{Keddis2015_Optimizingschedulesadaptable, Mayrhofer2020_CapabilityBasedProcess, Wenger2016_modelbasedengineering} \\
            \gls{bpmn}                   & \hspace{1sp}\citeS{Spitzer2020_genericApproachIndustrial} \\
            UML                          & \hspace{1sp}\citeS{Huang2021_AASCapabilityBased, Kathrein2019_EfficientProductionSystem} \\
            PackML                       & \hspace{1sp}\citeS{Heuss2020_IntegrationAutonomousTask}
        \end{tabularx}
        \label{tab:technologies}
    \end{table}
}

Considering UML, \gls{emf} or ontologies and even \gls{bpmn} are rather general frameworks used across several domains in software engineering. 
Both the OPC~UA Information Model and AutomationML were designed specifically for modeling automation and control systems, but are sometimes not flexible enough to apply more generic approaches. 
This could be the reason for a wide use of ontologies to create more holistic models that are not limited to the PLC/SCADA level.

Nevertheless, the modeling framework is mostly seen as tool for defining meta-models specifying the relationships between products, skills, capabilities and other assets. 
A significant number of rather abstract approaches rely on the \gls{ppr} model specifying capabilities as abstract definition of services provided, e.g., by machines, whereas skills are seen as executable programs or functions implementing these capabilities.

In particular, the approaches based on OPC~UA often do not offer the possibility to use capabilities and focus on modeling and implementing capabilities as ``callable'' services on an OPC~UA server. 
These callable services are often implemented as OPC~UA Programs or Methods. 
Those approaches that rely on more standardized frameworks such as VDI3682, VDI2206, ISA88, etc. focus mainly on creating holistic models on the one hand or suggesting very domain-specific models/exchange formats (e.g., PackML). In many cases specific guidance for implementation on standardized PLC/SCADA hardware is not provided or supported.

When summarizing the data collected, in most cases the meta-models are created based on the common standards or methodologies listed in Table~\ref{tab:metamodels}.
{
    \renewcommand{\arraystretch}{1.1}
    \begin{table}[ht]
        \caption{Meta-Models used for skill-based systems engineering.}
        \centering
        \begin{tabularx}{\columnwidth}{Xl}
            Meta-Model                         & Publication(s) \\
            \hline
            \gls{ppr} model and derivatives    & \hspace{1sp}\citeS{Motsch2021_ConceptModelingUsage, Zimmermann2021_ConceptSelectingSuitable, Herzog2020_AllocationPPRSplant, Meixner2020_ModelingExpertKnowledge, Sarna2021_ReducingRiskIndustrial, Evers2019_RoadmapSkillBased} \\
            VDI3682 Formal Process Description & \hspace{1sp}\citeS{Hoang2019_CapabilityModelAdaptation, Koecher2020_AutomatingDevelopmentMachine} \\
            VDI2206 Resource Definition        & \hspace{1sp}\citeS{Koecher2020_AutomatingDevelopmentMachine} \\
            VDMA Integrated Assembly Solutions & \hspace{1sp}\citeS{Zimmermann2019_SkillbasedEngineering} \\
            ISA88 State Machine Definition     & \hspace{1sp}\citeS{Koecher2020_AutomatingDevelopmentMachine}
        \end{tabularx}
        \label{tab:metamodels}
    \end{table}
}

Implementing these models and providing a means of execution the following technologies/runtime systems a lot of approaches do not provide specific information on how to execute these models within specific automation system, e.g., PLC or SCADA device. Those approaches relying on the OPC~UA Information Model are considered to be implementable on any OPC~UA capable device as long as the OPC~UA root structure is freely configurable. Relying on custom OPC~UA information models some approaches implement their own execution environment or make use of newer PLC standards such as IEC 61499 \citeS{Sonnleithner2021_IEC61499Distributed} and its mostly open source execution environments such as 4diac~\citeS{Motsch2021_ConceptModelingUsage, Dorofeev2021_GenerationOrchestratorCode, Dorofeev2019_EvaluatingSkillBased}.

Generally speaking, unfortunately, there appears to be no clear trend in specific implementation platforms, which could lead to an upcoming demand for standardized skill execution environments.

\subsection{Challenges for Capabilities \& Skills} 
\label{sec:Challenges}
\begin{figure}[ht]
    \centering
    \includegraphics[width=.425\textwidth]{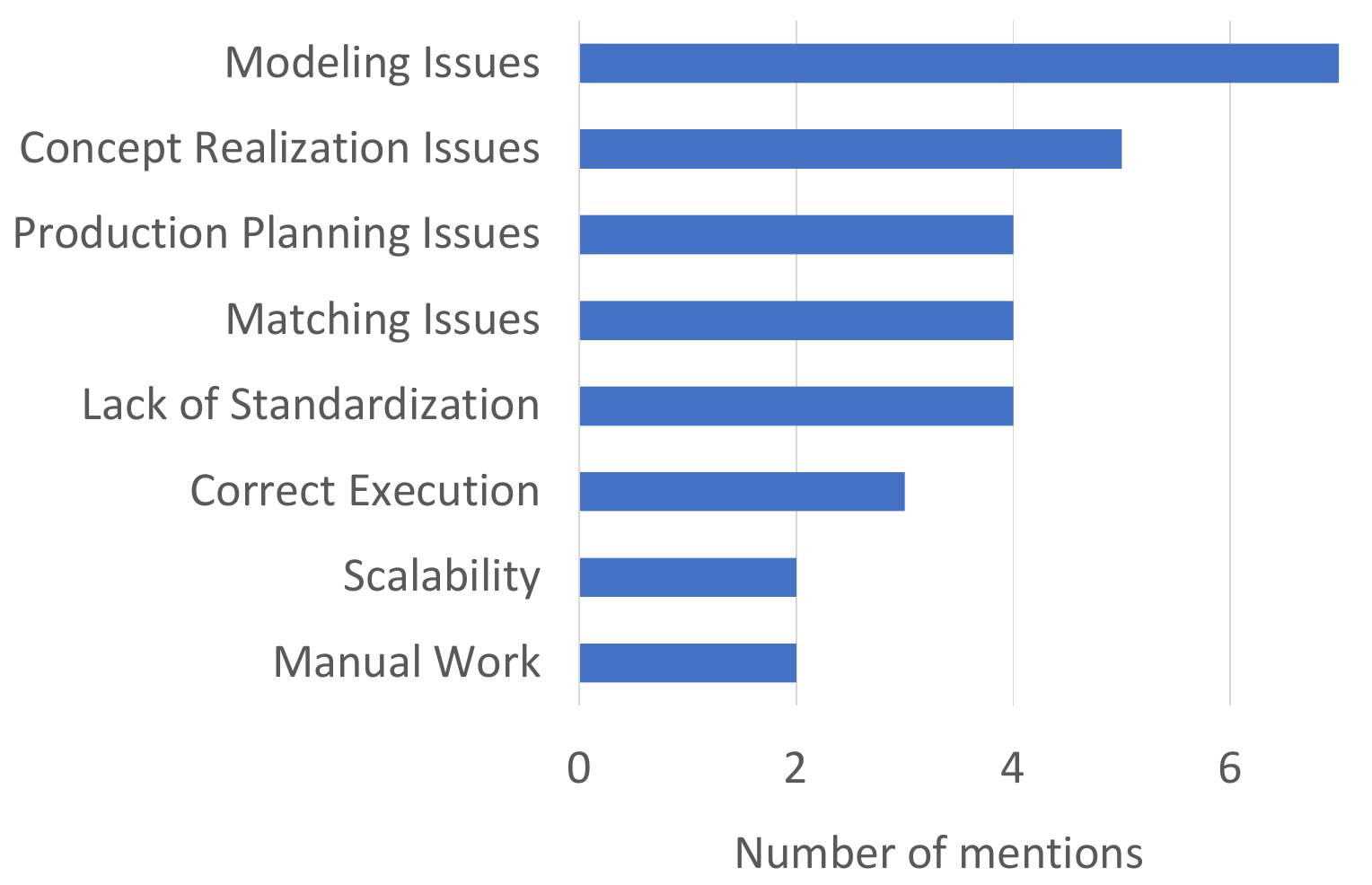}
    \caption{Overview of identified challenges.}
    \label{fig:challenges}
\end{figure} 

Out of 34 publications, 21 describe challenges that need to be solved to use capabilities \& skills in manufacturing.
We have condensed these challenges in Figure~\ref{fig:challenges}. 
\emph{Modeling Issues} are the challenge mentioned most often (e.g. \cite{Malakuti2018_ChallengesSkillbased,Eymueller2021_TowardsRealTime,Meixner2020_ModelingExpertKnowledge}).
For example, \cite{Malakuti2018_ChallengesSkillbased} noted different levels of abstraction in capability descriptions.
\cite{Meixner2020_ModelingExpertKnowledge, Koecher2020_AutomatingDevelopmentMachine} had additional modeling effort for the initial capabilities due to a lack of tools.
\cite{Koecher2020_FormalCapabilitySkill} faced incorrectly modeled capabilities that needed to be filtered out by consistency checking. 
The second most mentioned challenge \emph{Concept Realization Issues} summarizes issues that occurred while implementing capabilities \& skills. 
\cite{Profanter2019_HardwareAgnosticOPC} mention a lack of OPC UA support, especially for specific companion specifications, throughout the industry.
\cite{Spitzer2020_genericApproachIndustrial} expresses incompatibility problems with different OPC UA implementations and the discoverability of skills over different OPC UA servers. 
Furthermore, \cite{Dorofeev2019_AgileOperationalBehavior} and \cite{Soerensen2020_TowardsDigitalTwins} face implementation issues, like the detection of unplugged or shutdown devices and the need for extending hardware interfaces toward modern interfaces that enable real-time control of robots. 
Further, \emph{Production Planning Issues} occur, e.g., in \cite{Herzog2020_AllocationPPRSplant,Hoang2019_CapabilityModelAdaptation} and \cite{Zimmermann2021_ConceptSelectingSuitable}. \cite{Hoang2019_CapabilityModelAdaptation} miss automation support in production planning, and in \cite{Herzog2020_AllocationPPRSplant}, the generated production plans had to be checked and evaluated manually. 
\emph{Matching Issues} are a subclass of the production planning issues.
\cite{Hoang2019_CapabilityModelAdaptation} and \cite{Kathrein2019_EfficientProductionSystem} experience inefficient matching processes. 
\cite{Spitzer2020_genericApproachIndustrial} face issues regarding the skill identification and the naming of variables. 
The \emph{Lack of Standardization} of taxonomies, formal descriptions, and skill interfaces is mentioned in \cite{Malakuti2018_ChallengesSkillbased,Spitzer2021_ATLASGenericFramework} and \cite{Spitzer2020_genericApproachIndustrial}. 
Due to this gap, \cite{Huang2021_AASCapabilityBased} have issues regarding the interoperability of capability descriptions. 
The \emph{Correct Execution} of skills is also challenging. 
\cite{Pfrommer2013_PPRSProductionskills} mention that the skill execution is likely hindered by unforeseen events. 
\cite{Huang2021_AASCapabilityBased} require effective monitoring of the production lines to ensure the \emph{Correct Execution} of production plans. 
The authors in \cite{Spitzer2021_ATLASGenericFramework} had a tremendous effort to set up appropriate digital models in 3D simulation environments to monitor the \emph{Correct Execution}.
The last two challenges are \emph{Scalability} and \emph{Manual Work}. 
\cite{Evers2019_RoadmapSkillBased} reported issues due to the vast search space resulting from large product catalogs and a wide range of constraints. 
\cite{Himmelhuber2020_OntologyBasedSkill} mentioned the lack of digitized skill descriptions and their extraction skill descriptions and the manual work needed to describe capabilities \& skills.


\section{Discussion} 
\label{sec:Discussion}

From the analysis we condensed the following findings.
The data shows that the community is strongly based in Europe with a high industry influence and a growth of publications over the last three years (\emph{RQ1}).
Main requirements toward capabilities \& skills are a \emph{formal description}, the \emph{(re-)configurability}, \emph{matchability}, and \emph{statefulness} of skills (cf. Section~\ref{sec:Requirements}).
The community expects that industry benefits in \emph{flexibility} and \emph{planning efficiency} from using skills (cf. Section~\ref{sec:Benefits}).
While \emph{\gls{ppr}} models are most often the basis of capabilities, e.g., modelled using \emph{ontologies}, technologies currently used for skill implementations are \emph{OPC UA}, and \emph{AutomationML} (cf. Section~\ref{sec:Concepts}).

To investigate the understanding of capability and skill definitions (\emph{RQ2}), we compiled two analyses (cf. Section~\ref{sec:Definition}).
We provided a map of the publications with the notion of their capability and skill definitions that also reveals the dependencies between these definitions.
Furthermore, we categorized the definitions we identified into five groups and provide an overview of the definitions most used.

Current research on capabilities \& skills in manufacturing still has open gaps and opportunities for further research (\emph{RQ3}).
The data shows that many approaches are only \emph{scarcely evaluated} with mostly \emph{small examples} (cf. Section~\ref{sec:Usecase}).
Furthermore, the data shows that only a few publications provide data for their approach (cf. Section~\ref{sec:Usecase}).
This indicates that existing approaches should be examined with real-world use cases in a more reproducible and replicable way.
Challenges that need to be addressed before skills and capabilities can be used in practice are \emph{modeling issues} like different levels of abstraction as well as \emph{concept realization issues} with current technologies (cf. Section~\ref{sec:Challenges}).

\section{Conclusion} 
\label{sec:Conclusion}

An essential goal of Industry 4.0 is the adaptive production of mass-customizable products.
Capabilities \& skills are studied as an abstraction of production processes and resources to support this goal (cf. Section~\ref{sec:Motivation}).
The \gls{etfa} conference with the special session on capabilities \& skills provides a forum for academic and industrial research in this field.

To investigate this research area, we conducted a concise literature survey on \gls{etfa} publications from the last ten years.
We identified 247 publications that mention skills or capabilities in their metadata.
From these publications, we analyzed 34 papers that were within the focus of this work.

We provide an overview of the current state of research and categorized the definitions used in the literature including their links to prior publications.
Furthermore, we outlined remaining challenges, such as modeling and realization issues, and raised gaps in the evaluation and reproducibility of existing research.

\appendices

\section*{Acknowledgment}
The financial support by the Christian Doppler Research Association, the Austrian Federal Ministry for Digital and Economic Affairs and the National Foundation for Research, Technology and Development is gratefully acknowledged. 
This work is partly funded by the German Federal Ministry of Education and Research through the project ``BaSys 4.2'' (grant agreement no. 01IS19022).

\bibliographystyle{IEEEtran}
\bibliography{references} 

\bibliographystyleS{IEEEtran}
\bibliographyS{ETFA_2012-2021_Skill-Survey_v4} 

\end{document}